%% file: HD-MoGGE_V5.tex
\documentclass[runningheads]{llncs}

\usepackage{amsmath,amssymb, dsfont}
\usepackage{graphicx}
\usepackage{caption}
\usepackage{dcolumn}
\usepackage{bm}
\usepackage{float}
\usepackage{url}
\usepackage{subfig}
\usepackage{array}
\usepackage[table]{xcolor}
\usepackage{multicol}
\usepackage{multirow} 

\usepackage{hyperref}

\DeclareMathOperator{\sign}{sign}
\input{mymathsym.tex}

\begin{document}

\title{Regularized Estimation and Feature Selection in Mixtures of Gaussian-Gated Experts Models}

\titlerunning{Regularized Mixtures of Gaussian-Gated Experts Models}

\author{
Fa\"{i}cel Chamroukhi\inst{1} \and Florian Lecocq\inst{2} \and Hien D. Nguyen\inst{3}}

\authorrunning{Chamroukhi et al.}

\institute{
Department of Mathematics, University of Queensland\\ Brisbane, 4072 Queensland, Australia \and
University of Caen, Laboratory of Mathematics Nicolas Oresme \\  LMNO - UMR CNRS, Unicaen Campus 2, 14000 Caen, France
\and
Department of Mathematics and Statistics, La Trobe University \\ Melbourne Victoria 3086, Australia
}

\maketitle

\begin{abstract}
Mixtures-of-Experts models and their maximum likelihood estimation (MLE) via the EM algorithm have been thoroughly studied in the statistics and machine learning literature. They are subject of a growing investigation in the context of modeling with high-dimensional predictors with regularized MLE. We examine MoE with Gaussian gating network, for clustering and regression, and propose an $\ell_1$-regularized MLE to encourage sparse models and deal with the high-dimensional setting. We develop an EM-Lasso algorithm to perform parameter estimation and utilize a BIC-like criterion to select the model parameters, including the sparsity tuning hyperparameters. Experiments conducted on simulated data show the good performance of the proposed regularized MLE compared to the standard MLE with the EM algorithm.


\end{abstract}

\section{Introduction}

Mixture-of-experts  (MoE), originally introduced in~\cite{jacobsME,Jordan-HME-1994}, form a class of conditional mixture models \cite{McLachlan2000FMM} for modeling, clustering and prediction in the presence of heterogeneous data.  Their construction rely on conditional mixture models ~\cite{McLachlan2000FMM} in which both the gating network, formed by the mixing proportions, and the experts network formed by the mixture components, depend on the predictors or the inputs.  
The most popular choices for the gating network are the softmax gating functions \cite{jacobsME} or the Gaussian gating functions; the latter is a particular case of the exponential family gating functions introduced in \cite{XuJordan95}. 

Different choices are now common for the expert network model, depending on the type of the observed responses. For instance, a model for normal observations for regression and clustering was introduced in~\cite{Chamroukhi-NMoE-TimeSeries-IJCNN2009} or non-normally distributed expert models like in \cite{Chamroukhi-NNMoE-2015} to deal with skewed data distributions \cite{Chamroukhi-SNMoE-2016}, to ensure robustness to outliers \cite{Chamroukhi-TMoE-2016,Nguyen2016-MoLE}, or to accommodate both skewness and robustness as in \cite{Chamroukhi-STMoE-2017}. A detailed review on MoE models can be found in~\cite{NguyenChamroukhi-MoE}

Fitting MoE is generally performed by maximum-likelihood estimation (MLE) via the EM algorithm or its variants~\cite{dlr,McLachlanEM2008}. 
In a high-dimensional setting, the regularization of the MLE, to perform parameter estimation under a sparsity hypothesis and hence to simultaneously perform feature selection, has been studied in \cite{Khalili2010} and more recently in \cite{Chamroukhi-RMoE-2018,Huynh-Chamroukhi-prEMME-2019}. These approaches consider $\ell_1$ and $\ell_2$ penalties for the log-likelihood function, and are constructed upon softmax gating functions. 

In this paper, we consider MoE with Gaussian gated functions, and propose an $\ell_1$-regularized MLE via and EM-Lasso algorithm. We study the performance of the proposal on an experimental setup.
The remainder of this paper is organized as follows.
Section \ref{sec:GG-MoE} describes the MoE modeling framework, and the Gaussian-gated MoE and its MLE with the EM algorithm. Then, Section \ref{sec:PMLE-GGMoE} presents the proposed regularized MLE and the EM-Lasso algorithm. Finally, Section \ref{sec:Experiments} is dedicated to numerical experiments.

 \section{Gaussian-Gated Mixture-of-Experts}
\label{sec:GG-MoE}

\subsection{MoE modeling framework}
We consider mixtures-of-experts model to relate a high-dimensional predictor $\bsX\in \R^p$ to a response $\bsY \in \R^d$, potentially multivariate $d\geq 1$. We assume that the pair $(\bsX,\bsY)$ is generated from a heterogeneous population governed by a hidden structure represented by a latent categorical variable $Z\in [K]=\{1,\ldots,K\}$. Assume that we observe a random sample $\{(\bsX_i,\bsY_i)\}_{i=1,\ldots,n}$ of $n$ independently and identically distributed (i.i.d) pairs $(\bsX_i,\bsY_i)$ from $(\bsX,\bsY)$, and let $\cD = ((\bsx_1, \bsy_1),\ldots,(\bsx_n, \bsy_n))$ be an observed data sample. 
Assume that the pair $(\bsX,\bsY)$ follows a MoE distribution, then the MoE model can be defined as
\begin{equation}
    f(\bsy_i|\bsx_i;\bsvPsi) = \sum_{k=1}^{K}  g_{k}(\bsx_i;\bw) f(\bsy_{i}|\bsx_i;\bstheta_k)
    \label{eq:MoE}
\end{equation}where $g_{k}(\bsx;\bw) = \Pro(Z=k|\bsX=\bsx;\bw)$ is the distribution of the hidden variable $Z$ given the predictor $\bsx$ with parameters $\bw$, which represents the gating network, and the conditional component densities $f(\bsy|\bsx;\bstheta_k) = f(\bsy_{i}|\bsX=\bsx,Z=k;\bstheta)$ represent the experts network whose parameters are $\bstheta_k$.

\subsection{Gaussian-Gated Mixture-of-Experts}
\label{ssec:GG-MoE}

Let us define by $\phi_m(\bsv;\bsm,\bC) = (2\pi)^{-m/2}|\bC|^{-1/2}\exp\left(-\frac{1}{2} (\bsv-\bsm)^\top \bC^{-1} (\bsv-\bsm)\right)$ the probability density function of a Gaussian random vector $\bsV$ of dimension $m$ with mean $\bsm$ and covariance matrix $\bC$.
We consider mixture-of-experts for clustering and regression of heterogeneous data. In this case, the mixture of Gaussian-gated experts models, we abbreviate as MoGGE, for multivariate real responses, is defined by \eqref{eq:MoE} where the experts are (multivariate) Gaussian regressions, given by
\begin{equation}
    f(\bsy_{i}|\bsx_i;\bstheta_k) = \phi_d(\bsy_{i};\ba_{k} + \bB_{k}^T \bsx_{i},\bSigma_k)
    \label{eq:mvGaussian experts}
    \end{equation}
    and the gating network $g(\bsx_i;\bw)$ is defined by Gaussian gating function of the form:
\begin{equation}
    g_{k}(\bsx_i;\bw) = \frac{\Pro(Z_i=k) f(\bsx_{i}|Z_i=k;\bsw_k)}{\sum_{\ell=1}^{K}\Pro(Z_i=\ell) f(\bsx_{i}|Z_i=\ell;\bsw_{\ell})} = \frac{\alpha_k \phi_p(\bsx_{i};\bsmu_k,\bR_k)}{\sum_{\ell=1}^{K}\alpha_\ell \phi_p(\bsx_{i};\bsmu_\ell,\bR_\ell)}
\label{eq:Gaussian Gating Net}
\end{equation}with $\Pro(Z_i=k)=\alpha_k$, $f(\bsx_{i}|Z_i=k;\bw) = \phi_p(\bsx_{i};\bsmu_k,\bR_k)$ ($k=1,\ldots,K$). This Gaussian gating network was introduced in~\cite{XuJordan95} to sidestep the need for a nonlinear optimization routine in the inner loop of the EM algorithm in the case of a softmax function for the gating network.  
The MoGGE model is thus parameterized by the parameter vector 
 $\bsvPsi = (\bw^T,\bstheta^T)^T$ 
where 
$\bw = (\bsw^T_1,\ldots,\bsw^T_K)^T$ is the parameter vector of the gating network and 
$\bstheta = (\bstheta^T_1,\ldots,\bstheta^T_K)^T$ is the parameter vector of experts network, with $\bsw_k = (\alpha_k,\bsmu^T_k,{\text{vech}(\bR_k)^T})^T$ and $\bstheta_k = (\ba^T_k,\bB^T_k,{\text{vech}(\bSigma_k)^T})^T$ for $k = 1,\ldots,K$. 
The approximation capabilities of this model have been studied very recently in~\cite{Nguyen-Chamroukhi-Forbes-2019}.

\subsection{Maximum likelihood estimation via the EM algorithm}
Mixtures-of-experts of the form \eqref{eq:MoE} with softmax gating functions
are in general estimated by maximizing the (conditional) log-likelihood
$\sum_{i=1}^{n}\log f(\bsy_i|\bsx_i;\bsvPsi)$ by using the EM algorithm , in which the M-step requires an internal iterative numerical optimization procedure (eg. a Newton-Raphson algorithm)  to update the softmax parameters. 
We follow the approach of estimating MoGGE in \cite{XuJordan95}, which relies on maximizing the joint loglikelihood, and in the MLE, the M-Step can then be solved in a closed form. 
Indeed, based on equations~\eqref{eq:MoE}, \eqref{eq:mvGaussian experts}, and \eqref{eq:Gaussian Gating Net}, we the MoGGE conditional density is given by:
\begin{equation}
    f(\bsy_i|\bsx_i;\bsvPsi) = \sum_{k=1}^{K}  \frac{\alpha_k \phi_p(\bsx_{i};\bsmu_k,\bR_k)}{\sum_{\ell=1}^{K}\alpha_\ell \phi_p(\bsx_{i};\bsmu_\ell,\bR_\ell)} \phi_d(\bsy_{i};\ba_{k} + \bB_{k}^T \bsx_{i},\bSigma_k)\cdot
\end{equation}Then we can write the joint density as:
\begin{eqnarray} 
    f(\bsy_i,\bsx_i;\bsvPsi) &=&  f(\bsx_i;\bw)f(\bsy_i|\bsx_i;\bstheta)= \sum_{k=1}^{K} \Pro(Z_i=k)f(\bsx_i;\bsw_k)f(\bsy_i|\bsx_i;\bstheta_k)\nonumber\\
    &=& \sum_{k=1}^{K} \alpha_k \phi_p(\bsx_{i};\bsmu_k,\bR_k)\phi_d(\bsy_{i};\ba_{k} + \bB_{k}^T \bsx_{i},\bSigma_k)\cdot
    \label{eq:joint density}
\end{eqnarray}The joint log-likelihood to be maximized by EM is therefore given by:
\begin{equation}\label{eq:joint loglik}
L(\bsvPsi) = \sum_{i=1}^{n}\log f(\bsy_i,\bsx_i;\bsvPsi) = \sum_{i=1}^{n}\log \sum_{k=1}^{K} \alpha_k \phi_p(\bsx_{i};\bsmu_k,\bR_k)\phi_d(\bsy_{i};\ba_{k} + \bB_{k}^T \bsx_{i},\bSigma_k)\cdot
\end{equation}

\subsection{The EM algorithm for the MoGGE model}
 
The complete-data log-likelihood upon which the EM principle is constructed is then defined by
\begin{equation}
    L_c(\bsvPsi) = \sum_{i=1}^{n}\sum_{k=1}^{K} Z_{ik} \log \left[\alpha_k \phi_p(\bsx_{i};\bsmu_k,\bR_k)\phi_d(\bsy_{i};\ba_{k} + \bB_{k}^T \bsx_{i},\bSigma_k) \right]
    \label{eq:complete joint log-lik}
\end{equation}where $Z_{ik}$ being an indicator binary-valued variable such that $Z_{ik}=1$ if $Z_i=k$ (i.e., if the $i$th pair $(\bsx_i,\bsy_i)$ is generated from the $k$th expert and $Z_{ik}=0$ otherwise.
The EM algorithm, after starting with an initial solution $\bsvPsi^{(0)}$, alternates between the E- and the M- Steps until convergence (when there is no longer a significant change in the log-likelihood (\ref{eq:joint loglik})).
 \paragraph{E-step:}
\label{ssec: E-step}
Compute the expectation of the complete-data log-likelihood (\ref{eq:complete joint log-lik}), given the observed data $\cD$ and the current parameter vector estimate $\bsvPsi^{(q)}$: 
\begin{eqnarray}
    Q(\bsvPsi;\bsvPsi^{(q)}) &=&  \E\left[L_c(\bsvPsi)|\cD;\bsvPsi^{(q)}\right]\nonumber\\
    &=& \sum_{i=1}^{n}\sum_{k=1}^{K}\tau_{ik}^{(q)} \log \left[\alpha_k \phi_p(\bsx_{i};\bsmu_k,\bR_k)\phi_d(\bsy_{i};\ba_{k} + \bB_{k}^T \bsx_{i},\bSigma_k)\right],
    \label{eq:Q-function}
\end{eqnarray}where:
\begin{equation}
    \tau_{ik}^{(q)} = \Pro(Z_i=k|\bsy_i,\bsx_i;\bsvPsi^{(q)}) = \frac{ \alpha^{(q)}_k \phi_p(\bsx_{i};\bsmu_k^{(q)},\bR_k^{(q)}) \phi_d(\bsy_{i};\ba^{(q)}_{k} + {\bB^{(q)}_{k}}^T \bsx_{i},\bSigma^{(q)}_k)}{f(\bsx_i,\bsy_{i};\bsvPsi^{(q)})},
\label{eq:post prob}
\end{equation}
is the posterior probability that the observed pair $(\bsx_i, \bsy_i)$ is generated by  the $k$th expert. This step therefore only requires  the computation of the posterior  component membership probabilities $\tau^{(q)}_{ik}$ $(i=1,\ldots,n)$, for $k=1,\ldots,K$.

\paragraph{M-step:}
\label{ssec: M-step EM}
Calculate the parameter vector update $\bsvPsi^{(q+1)}$ by maximizing the $Q$-function (\ref{eq:Q-function}), i.e, 
$\bsvPsi^{(q+1)} = \arg \max_{\bsvPsi} Q(\bsvPsi;\bsvPsi^{(q)})$.
By decomposing the $Q-$function \eqref{eq:Q-function} as 
\begin{equation}\label{eq:Q-function decomposition}
Q(\bsvPsi;\bsvPsi^{(q)}) =\sum_{k=1}^{K}Q(\bsw_k;\bsvPsi^{(q)})+Q(\bstheta_k;\bsvPsi^{(q)})
\end{equation}where
\begin{equation}\label{eq:Q-function gating-network}
    Q(\bsw_k;\bsvPsi^{(q)}) =  \sum_{i=1}^{n}\tau_{ik}^{(q)} \log \left[\alpha_k \phi_p(\bsx_{i};\bsmu_k,\bR_k)\right]
\end{equation}
and
\begin{equation}\label{eq:Q-function expert-network}
    Q(\bstheta_k;\bsvPsi^{(q)}) = \sum_{i=1}^{n} \tau_{ik}^{(q)} \log \phi_d(\bsy_{i};\ba_{k} + \bB_{k}^T \bsx_{i},\bSigma_k), 
\end{equation}the maximization can then be done by performing  $K$ separate maximizations w.r.t the gating network parameters and the experts network parameters.

\paragraph{Updating the the gating networks' parameters:} 
Maximizing~\eqref{eq:Q-function gating-network} w.r.t $\bsw_k$'s corresponds to the M-Step of a Gaussian Mixture Model \cite{McLachlan2000FMM}. The closed-form expressions for updating the parameters are given by:
\begin{eqnarray}
    \alpha_{k}^{(q+1)} &=&\sum_{i=1}^{n}\tau_{ik}^{(q)}\Big/n,
    \label{eq:MLE alphak}\\
    \bsmu_{k}^{(q+1)} &=& \sum_{i=1}^{n}\tau_{ik}^{(q)}\bsx_i\Big/\sum_{i=1}^{n}\tau_{ik}^{(q)},
    \label{eq:MLE muk}\\
    \bR_{k}^{(q+1)} &=& \sum_{i=1}^{n}\tau_{ik}^{(q)}(\bsx_i-\bsmu_k^{(q+1)})(\bsx_i-\bsmu_k^{(q+1)})^T\Big/\sum_{i=1}^{n}\tau_{ik}^{(q)}\cdot
    \label{eq:MLE Rk}
\end{eqnarray}

 \paragraph{Updating the experts' network parameters}
Maximizing~\eqref{eq:Q-function expert-network} w.r.t $\bstheta_k$'s corresponds to the M-Step of standard MoE with multivariate Gaussian regression experts, see e.g \cite{Chamroukhi-MRHLP-2013}. 
The closed-form updating formulas are given by: 
{\small \begin{eqnarray}
\!\!\!\!\!\!\!\! \ba^{(q+1)}_{k} &\!\! =\!\! &  \sum_{i=1}^{n}  \tau_{ik}^{(q)}(\bsy_i - {\bB^{(q)}_{k}}^T\bsx_{i})\Big/ \sum_{i=1}^n \tau_{ik}^{(q)},
    \label{eq:MLE a_k}\\
\!\!\!\!\!\!\!\!    \bB^{(q+1)}_{k} &\!\! =\!\! &   \Big[\sum_{i=1}^{n}\tau^{(q)}_{ik} \bsx_i\bsx_i^T \Big]^{-1} \sum_{i=1}^{n}\tau^{(q)}_{ik} \bsx_i (\bsy_{i} - \ba^{(q+1)}_{k})^T,
    \label{eq:MLE Bk} \\
\!\!\!\!\!\!\!\!    \bSigma_{k}^{(q+1)} &\!\! =\!\! &  \sum_{i=1}^{n}  \tau_{ik}^{(q)}
    (\bsy_i -  (\ba^{(q+1)}_{k} + {\bB^{(q+1)}_{k}}^T\bsx_{i}))
    (\bsy_i -  (\ba^{(q+1)}_{k} + {\bB^{(q+1)}_{k}}^T\bsx_{i}))^T
    \!\!\Big/\!\!\sum_{i=1}^n \tau_{ik}^{(q)}\cdot
    \label{eq:MLE Sigmak}
\end{eqnarray}}
However, in a high dimensional setting, MLE may be unstable or even unfeasible. One possible way to proceed in such a context is the regularization of the objective function. 
In the context of MoE models, this has been studied namely in \cite{Khalili2010,Chamroukhi-RMoE-2018,Huynh-Chamroukhi-prEMME-2019} where  $\ell_1$ and $\ell_2$ regularization for the log-likelihood function of the standard MoE model with softmax gating network. This  penalized MLE allow an efficient estimation for simultaneous parameter estimation and feature selection.

\section{Penalized maximum likelihood parameter estimation}
\label{sec:PMLE-GGMoE}
Here we study the regularized estimation of the MoGGE model. We first consider the case when $d=1$ (univariate response $\bsy_i$). The expert densities are thus defined by
$f(y_{i}|\bsx_i;\bstheta_k) = \phi(y_{i};\beta_{k,0} + \bsbeta^T_k \bsx_i,\sigma^2_k)$ 
with  $\bstheta_k = (\beta_{k,0},\bsbeta^T_k,\sigma^2_k)^T$.

In our proposed approach, rather than maximizing the joint log-likelihood
\eqref{eq:joint loglik}, we attempt to maximize its $\ell_1$-regularized version, to encourage sparse models and to perform estimation and feature selection. The resulting penalized log-likelihood can then be defined by:
\begin{equation}
    \cL(\bsvPsi) = L(\bsvPsi) - \text{Pen}_{\lambda,\gamma}(\bsvPsi)
    \label{eq:pen joint loglik}
\end{equation}
where $L(\bsvPsi)$ is the observed-data log-likelihood of $\bsvPsi$ defined by~\eqref{eq:joint loglik} 
and $\text{Pen}_{\lambda,\gamma}(\bsvPsi)$ is a Lasso~\cite{Tibshirani96Lasso} regularization term encouraging sparsity for the expert network parameters  and the gating network parameters, with $\lambda$ and $\gamma$ positive real values representing tuning hyperparameters.
For regularizing the expert parameters, the penalty is naturally applied to the regression coefficient vectors $\bsbeta_k$.
For the gating network, since the estimates are those of a Gaussian mixture, 
we then follow the strategy of feature selection in model-based clustering in~\cite{Pan:2007:PenMBC} in which we apply the penalty to the Gaussian mean vectors $\bsmu_k$ and assume that the Gaussian covariance matrices of the gating network are diagonal, ie. $\bR_k = \text{diag}(\nu^2_1,\ldots,\nu_K^2)$.  
The penalty function is then given by:
\begin{equation}
    \text{Pen}_{\lambda,\gamma}(\bsvPsi) = \lambda \sum_{k=1}^{K}\Vert \bsbeta_k\Vert_1 + \gamma \sum_{k=1}^{K}\Vert \bsmu_k\Vert_1\cdot
\end{equation}We now derive an EM-Lasso algorithm to maximize \eqref{eq:pen joint loglik}.

\subsection{The EM-Lasso algorithm for the MoGGE model}
Lets first define the penalized joint complete-data log-likelihood, which is given by
\begin{equation}
    \cL_c(\bsvPsi) = L_c(\bsvPsi) - \text{Pen}_{\lambda,\gamma}(\bsvPsi)
    \label{eq:pen complete joint log-lik}
\end{equation}where $L_c(\bsvPsi)$ is the non-regularized joint complete-data log-likelihood defined by~\eqref{eq:complete joint log-lik}.
The EM-Lasso algorithm then alternates between the two following steps until convergence (when there is no significant change in \eqref{eq:pen joint loglik}.

\paragraph{E-step.}
\label{ssec: E-step EM}
This step computes the expectation of the complete-data log-likelihood (\ref{eq:pen complete joint log-lik}),  given the observed data $\cD$, using the current parameter vector $\bsvPsi^{(q)}$:
\begin{eqnarray}
    \mathcal{Q}_{\lambda,\gamma}(\bsvPsi;\bsvPsi^{(q)}) &=&  \E\left[\cL_c(\bsvPsi)|\cD;\bsvPsi^{(q)}\right] = Q(\bsvPsi;\bsvPsi^{(q)}) - \text{Pen}_{\lambda,\gamma}(\bsvPsi)
    \label{eq:Q-function penalized}
\end{eqnarray} 
which only requires the computation of the posterior probabilities of component membership $\tau^{(q)}_{ik}$ $(i=1,\ldots,n)$, for each of the $K$ experts as defined by \eqref{eq:post prob}.

\paragraph{M-step.}
\label{ssec: M-step EM penalized} 
This step updates the value of the parameter vector $\bsvPsi$ by maximizing the $Q$-function (\ref{eq:Q-function}) with respect to $\bsvPsi$, that is, by computing the parameter vector update $\bsvPsi^{(q+1)} = \arg \max_{\bsvPsi} \mathcal{Q}_{\lambda,\gamma}(\bsvPsi;\bsvPsi^{(q)}).$ 
Now we have this decomposition
\begin{eqnarray}
 \mathcal{Q}_{\lambda,\gamma}(\bsvPsi;\bsvPsi^{(q)}) &=&  \sum_{k=1}^K \mathcal{Q}_{\gamma}(\bsw_k;\bsvPsi^{(q)})+\mathcal{Q}_{\lambda}(\bsvPsi_k;\bsvPsi^{(q)})
\label{eq:Q-function decomposition penalized}
\end{eqnarray}and the maximization is performed by $K$ separate maximizations of the penalized $Q$-functions $\mathcal{Q}_{\gamma}(\bsw_k;\bsvPsi^{(q)})$ and $\mathcal{Q}_{\lambda}(\bsvPsi_k;\bsvPsi^{(q)})$. 

\paragraph{Coordinate Ascent for updating the gating network} 
Updating the gating network parameters consists of maximizing w.r.t $\bsw_k$ the following penalized $Q$-function
\begin{eqnarray}\label{eq:Q-function gating-network penalized}
    \mathcal{Q}_{\gamma}(\bsw_k;\bsvPsi) &=& \sum_{i=1}^{n}\tau_{ik}^{(q)} \log \left[\alpha_k \phi_p(\bsx_{i};\bsmu_k,\bR_k)\right] -\gamma \sum_{j = 1}^{p} |\mu_{k,j}| \nonumber\\
    &=& \sum_{i=1}^{n}\tau_{ik}^{(q)} \log \alpha_k + \sum_{i=1}^{n}\tau_{ik}^{(q)} \log\phi_p(\bsx_{i};\bsmu_k,\bR_k) -\gamma \sum_{j = 1}^{p} |\mu_{k,j}|\cdot \nonumber
\end{eqnarray}It can be seen that the updates of the $\alpha_k$'s are unchanged compared to the standard algorithm and are given by \eqref{eq:MLE alphak}. 
For the mean vectors,  updating the coefficients $\mu_{k,j}$ corresponds
to weighted version or and $\ell_1$-regularized maximum likelihood estimation a Gaussian mean; The coefficients $\mu_{k,j}$ can then be updated in a cyclic way by using a Coordinate ascent algorithm until \eqref{eq:Q-function gating-network penalized} is maximized. 
Coordinate ascent (CA) \cite[sec. 5.4]{TH15c} \cite{Friedman07pathwisecoordinate,wu2008} is indeed an efficient way to solve Lasso-regularization problems.
For each coefficient index $j=1,\ldots,p$, 
it can be easily shown that, after starting with the previous EM-Lasso estimate
as initial value, i.e, $\mu_{kj}^{(0, q)}=\mu_{kj}^{(q)}$, each iteration $t$ of the CA algorithm updates are given by the following updating formulas (see eg. \cite{Pan:2007:PenMBC}), written in a scalar and a vector form: 
\begin{eqnarray}
    \mu_{kj}^{(t+1, q)} &=& \sign(\tilde{\mu}_{kj}^{(q+1)}) \left( |\tilde{\mu}_{kj}^{(q+1)}| - \frac{\gamma}{\sum_{i=1}^{n}\tau_{ik}^{(q)}} \nu_{kj}^{2(q)} \right)_{+}\nonumber \\
    &=& \mathcal{S}\left(\sum_{i=1}^{n}\tau_{ik}^{(q)}x_{ij}; \gamma\nu_{kj}^{2(q)}\right)\Big/\sum_{i=1}^{n}\tau_{ik}^{(q)}\nonumber \\
&=& \mathcal{S}\left(\bX^T_j \bstau_k^{(q)}; \gamma\nu_{kj}^{2(q)}\right)/\Bs 1^T_n\bstau_k^{(q)}
\end{eqnarray}with,
$\tilde{\mu}_{kj}^{(q+1)} = \sum_{i=1}^{n}\tau_{ik}^{(q)}x_{ij}\Big/\sum_{i=1}^{n}\tau_{ik}^{(q)}$ is the usual non-regularized MLE update for $\mu_{k}$ 
(Eq.~\eqref{eq:MLE muk}), 
$\bX_j$ the $j$th column of $\bX$, 
$\Bs 1_n$ is a vector of ones of size $n$,
$\bstau^{(q)}_k = (\tau^{(q)}_{1k},\ldots,\tau^{(q)}_{nk})^T$, 
and
$\mathcal{S}(u;\eta):= \sign(u)(|u| - \eta)_+$ is the soft-thresholding operator with $(.)_+=\max\{., 0\}$. 
The CA procedure is iterated until no significant change in \eqref{eq:Q-function gating-network penalized} is observed. We then take the update at convergence of the CA algorithm, i.e $\mu_{kj}^{(q+1)} = \mu_{kj}^{(t+1, q)}$. 
Finally, the updates of the diagonal elements of the co-variance matrices are given by: \begin{equation}
\nu_{kj}^{2(q+1)} = \sum_{i=1}^{n}\tau_{ik}^{(q)}(x_{ij}-\mu_{kj}^{(q+1)})^2\Big/\sum_{i=1}^{n}\tau_{ik}^{(q)}
\cdot
\end{equation}

\paragraph{Coordinate Ascent for updating the experts network}
The maximization step for updating the expert parameters $\bstheta_k$ consists of maximizing the function $\mathcal{Q}_{\lambda}(\bstheta_k;\bsvPsi^{(q)})$ given by:
\begin{eqnarray*}
    \mathcal{Q}_{\lambda}(\bstheta_k;\bsvPsi^{(q)}) &=& Q(\bsvPsi_k;\bsvPsi^{(q)}) - \lambda \sum_{j = 1}^{p} |\beta_{k,j}|\nonumber\\
    &=& - \frac{1}{2\sigma_k^2}\sum_{i=1}^{n} \tau_{ik}^{(q)} \left(y_{i} - (\beta_{k,0} + \bsbeta_{k}^T \bsx_{i})\right)^2 - \frac{n_{k}^{(q)}}{2}\log(2\pi \sigma_k^2)- \lambda \sum_{j = 1}^{p} |\beta_{k,j}|\cdot \nonumber
    \label{eq:Q-function expert-network penalized}
\end{eqnarray*}
Updating $\bsbeta_{k}$, for each component $k$, consists of solving an independent weighted Lasso problem where the weights are the posterior component membership probabilities $\tau^{(q)}_{ik}$. Each of these weighted Lasso problems is then separately solved by Coordinate Ascent. 
The CA algorithm, after starting from the previous EM-Lasso estimate as initial values, i.e $\beta_{kj}^{(0,q)}=\beta_{kj}^{(q)}$, calculates, at each iteration $t$, the following coordinate updates, 
until no significant change in \eqref{eq:Q-function expert-network penalized}: 
\begin{eqnarray}
    \beta_{kj}^{(t+1,q)} 
    &=& \mathcal{S}\left(\sum_{i=1}^n\tau_{ik}^{(q)}r_{ikj}^{(t,q)}x_{ij};\lambda \sigma_k^{(q)2}\right)\Big/\sum_{i=1}^n\tau_{ik}^{(q)}x_{ij}^2\\
    &=& \mathcal{S}\left(\bX^T_j \bW_k^{(q)}\bsr_{kj}^{(q)}; \lambda{\sigma_k^{(q)}}^2\right)/(\bX^T_j\bW_k^{(q)}\bX_j),
\end{eqnarray}with $r_{ikj}^{(t,q)} = y_i-\beta_{k0}^{(q)}- \bsx_i^T\bsbeta^{(t,q)}_k  + \beta_{kj}^{(t,q)}x_{ij}$,  
$\bsr_{kj}^{(t,q)} = \bsy-\beta_{k0}^{(q)}{\Bs 1}_n - \bX \bsbeta^{(t,q)}_k + \beta_{kj}^{(t,q)}\bX_{j}$ is the residual without considering the contribution of the $j$-th coefficient, and
$\bW_k^{(q)} = \text{diag}(\Bs \tau_k^{(q)})$. 
The parameter vector update is then taken at convergence of the CA algorithm, i.e $\bsbeta^{(q+1)}_{k} = \bsbeta^{(t+1,q)}_{k}$. Then, the intercept and the variance, have the following standard updates: 
\begin{eqnarray}   
\!\!\!\! \beta^{(q+1)}_{k,0} &=& \sum_{i=1}^{n}  \tau_{ik}^{(q)}(y_i - \bsx_{i}^T\bsbeta^{(q+1)}_{k})\Big/ \sum_{i=1}^n \tau_{ik}^{(q)} = {\bstau_k^{(q)}}^T(\bsy - \bX \bsbeta^{(q+1)}_k)\big /\Bs 1_n^T \bstau_k^{(q)}
    \label{eq:PMLE betak_0}\\
\!\!\!\!    \sigma_{k}^{2(q+1)} &=& \sum_{i=1}^{n}  \tau_{ik}^{(q)}\left(y_i -  (\beta^{(q+1)}_{k,0} + \bsx_{i}^T\bsbeta^{(q+1)}_{k})\right)^2\Big/ \sum_{i=1}^n \tau_{ik}^{(q)}\\
    &=&
     \big \Vert\!\sqrt{\bW_k^{(q)}}\left(\bsy - \bsbeta_{k,0}^{(q+1)}{\Bs 1}_n - \bX \bsbeta^{(q+1)}_{k}\right)\!\big \Vert_2^2 \big /\Bs 1_n^T \bstau_k^{(q)}\cdot 
    \label{eq:PMLE sigmak2}
\end{eqnarray}

\subsection{Algorithm tuning and model selection}

In practice, appropriate values of the tuning parameters $(\lambda, \gamma)$ as well as the number of experts $K$ should be chosen.  In order to select them, we use a modified BIC based on a grid of candidate values for $K$, $\lambda$ and $\gamma$. 
This modified BIC is an extension of the criterion used in~\cite{Stadler2010} for regularized mixture of regressions and was used in~\cite{Chamroukhi-RMoE-2018,Huynh-Chamroukhi-prEMME-2019} and is defined as:
\begin{equation}\label{eq:modified BIC}
    \text{BIC}(K, \lambda, \gamma) = L(\widehat{\bsvPsi}_{K, \lambda, \gamma}) - \text{df}(K, \lambda, \gamma)\frac{\log n}{2},
\end{equation}where $\widehat{\bsvPsi}_{K, \lambda, \gamma}$ is the penalized log-likelihood estimator obtained by the EM-Lasso algorithm, and $\text{df}(K,\lambda, \gamma)$ is the estimated number of non-zero coefficients in the model, interpreted as the degrees of freedom. Let's assume that $K_{0} \in \{K_{1},\ldots,K_{M}\}$, whith $K_{0}$ the true number of expert components. For each value of $K$, we define grids of tuning parameters $\{\lambda_{1},\ldots,\lambda_{M_{1}}\}$ and $\{\gamma_{1},\ldots,\gamma_{M_{2}}\}$. For each triplet $(K, \lambda, \gamma)$, we calculated the penalized log-likelihood estimators $\widehat{\bsvPsi}_{K, \lambda, \gamma}$ and compute $\text{BIC}(K, \lambda, \gamma)$.
Finally, the model with parameters $(K, \lambda, \gamma)$ having the highest BIC value, is then selected.

\section{Experimental study}
\label{sec:Experiments}
In this section, we study the performance of our approach on simulated data. 
The codes are written in Matlab and in R and will be made publicly available on \url{https://github.com/fchamroukhi}. 
Different evaluation criteria are used to assess the model's performance, including sparsity, estimation of parameters and clustering accuracy.

\paragraph{Sparsity performance} In order to evaluate the sparsity of the model, we calculate the specificity/sensitivity defined by:
\begin{itemize}
    \item \textsf{Sensitivity:} proportion of correctly estimated zero coefficients;
    \item \textsf{Specificity:} proportion of correctly estimated nonzero coefficients.
\end{itemize}
\paragraph{Clustering performance} For measuring the clustering performance, we calculate the correct classification rate and the Adjusted Rate index (ARI) between the true simulated partition and the partition estimated by the EM algorithms.
The estimated cluster labels are obtained by plugin the Baye's allocation rule for the estimated model, which consists of maximizing the posterior probabilities defined in~\ref{eq:post prob} and calculated with the estimated parameters.
That is, the estimated class label $\hat{z}_{i}$ for the $i$-th pair $(\bsX_i,\bsY_i)$ is given by
\begin{equation}
    \hat{z}_{i} = \arg \max_{k=1}^K \tau_{ik}(\hat{\bsvPsi}) \quad (i=1,\ldots, n)\cdot 
\end{equation}For calculating the classification rate, we evaluate all the possible permutations of the obtained partition, and the one giving the best rate is then retained.  

\subsection{Simulation study}

The data are generated according to the following generative hierarchical process:
\begin{eqnarray*}
    Z_i &\sim& \text{Mult} (1;\alpha_1,\ldots,\alpha_K)\\
    \bsX_i|Z_i=z_i &\sim &\cN_p (.;\bsmu_{z_i},\bR_{z_i})\\
    \bsY_i|\bsX_i=\bsx_i, Z_i=z_i &\sim& \cN_d (.;\beta_{z_i,0} + \bsbeta_{z_i}^T \bsx_{i},\sigma^2_{z_i}).
\end{eqnarray*}We consider a MoGGE model of $K = 2$ expert components. 
The parameters of the Gaussian gating function, whose prior probabilities are  $\alpha_1 = \alpha_2 = 0.5$, are 
$\bsmu_1 = (0, 1, -1, -1.5, 0, 0.5, 0, 0)^T$, $\bsmu_2 =(2, 0, 1, -1.5, 0, -0.5, 0, 0)^T$
and $\bR_1 = \bR_2 = \text{diag}(\nu_1^2,\ldots,\nu_K^2)$ with $\nu_1^2 = \ldots = \nu_K^2 = 1$.
The parameters of the Gaussian expert regressors are $\bsbeta_1 = (0, 1.5, 0, 0, 0, 1, 0, -0.5)^T$, $\bsbeta_2 = (1, -1.5, 0, 0, 2, 0, 0, 0.5)$,  and $\sigma_1 = \sigma_2 = 1$.
For each data set, we sample $n = 300$ data pairs, and for each experiment, $100$ datasets were generated to average the results and provide error bars. 
In order to get the best model for each sample in the sense of the BIC criterion, we estimated the penalized model with the following grids of values for the parameters: $\lambda = (0,1,2,\ldots,25)$, $\gamma = (0,1,2,\ldots,25)$; The minimum and maximum values selected for $\lambda$ and $\gamma$ are respectively $4$, $20$ and $3$, $18$. 
Then we selected the penalized model which maximizes the modified BIC value \eqref{eq:modified BIC}.
 The results will be provided in the parts below.

\subsubsection{Obtained results}
\begin{figure}[!h]
    \centering
    \includegraphics[width=0.42\textwidth]{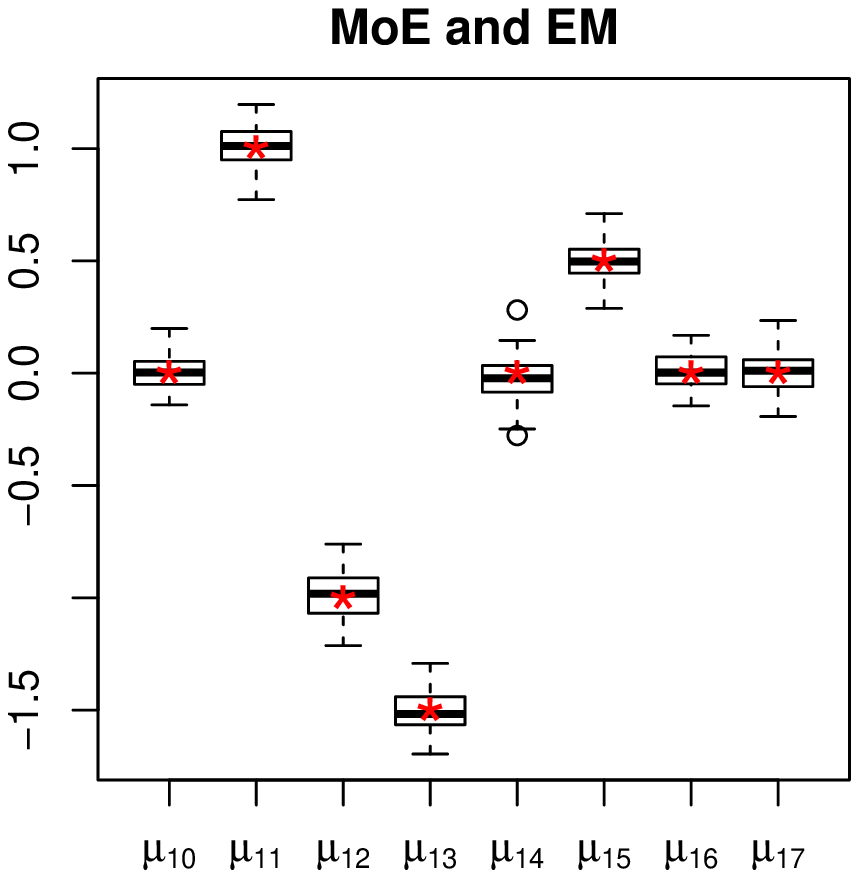}
    \includegraphics[width=0.42\textwidth]{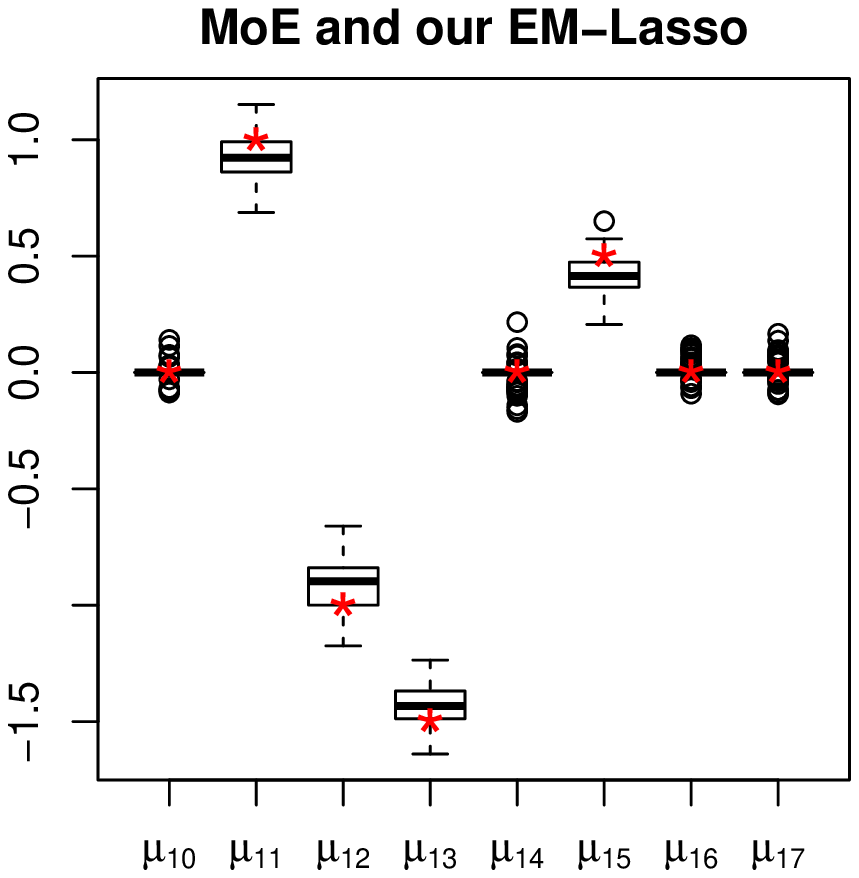}\\
    \includegraphics[width=0.42\textwidth]{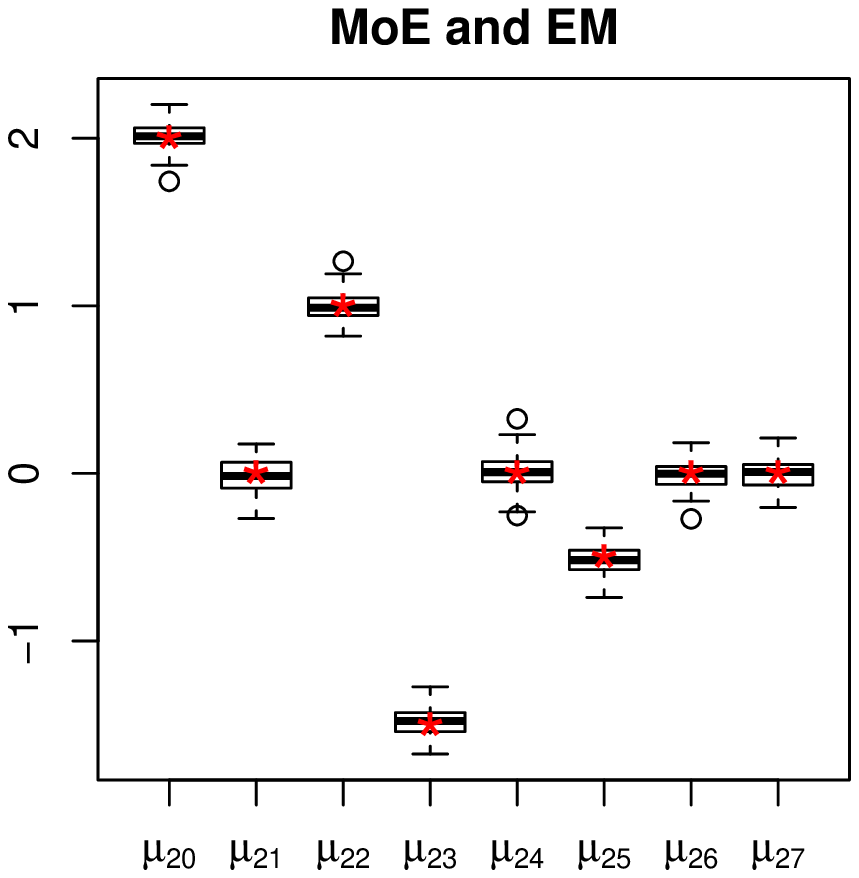}
    \includegraphics[width=0.42\textwidth]{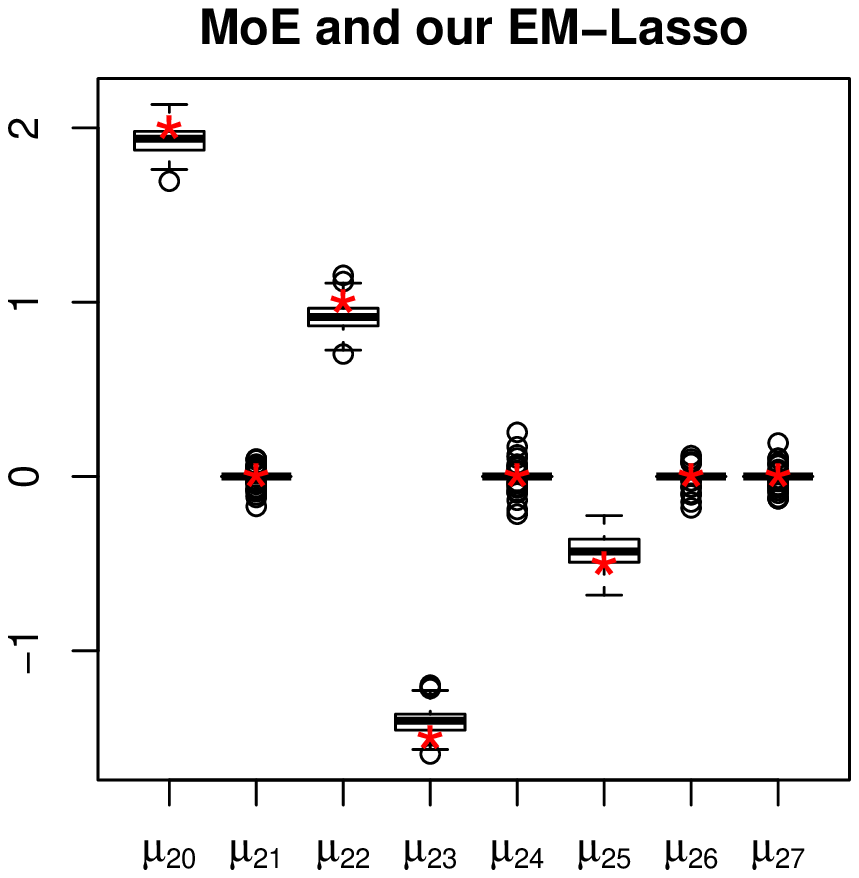}
    \caption{\label{fig:boxplot_gates}Boxplots of the estimated gating network parameters $\mu_{k,j}$: component $k=1$, top, and component $k=2$, bottom. The red stars are the true values.} 
\end{figure}

\begin{figure}[!h]
    \centering
    \includegraphics[width=0.42\textwidth]{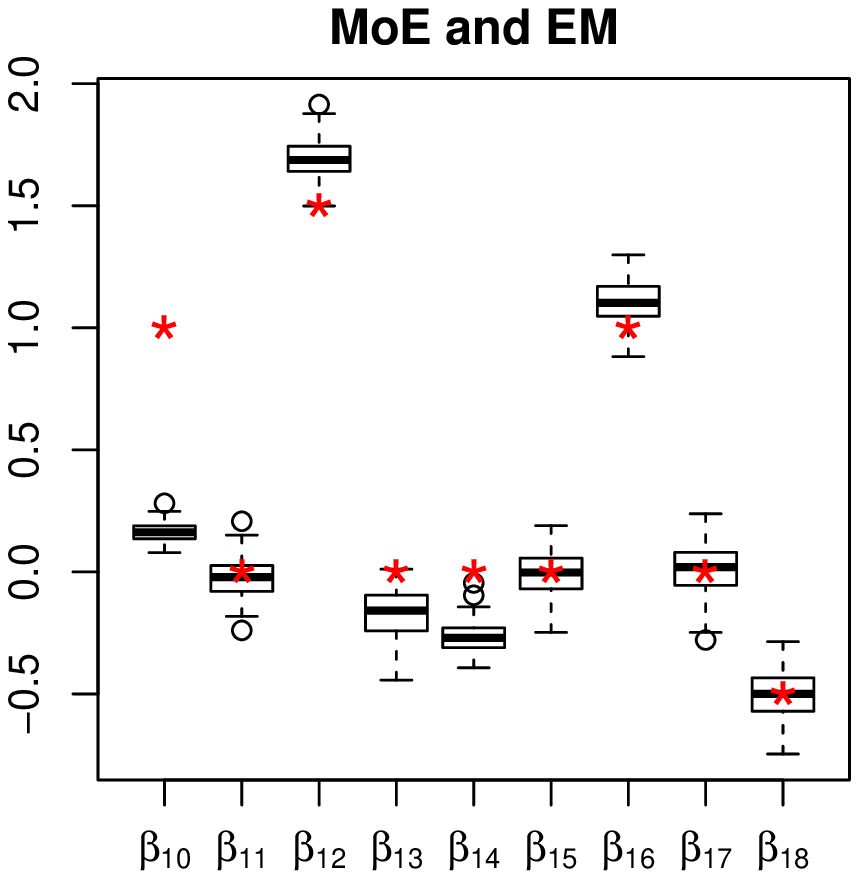}
    \includegraphics[width=0.42\textwidth]{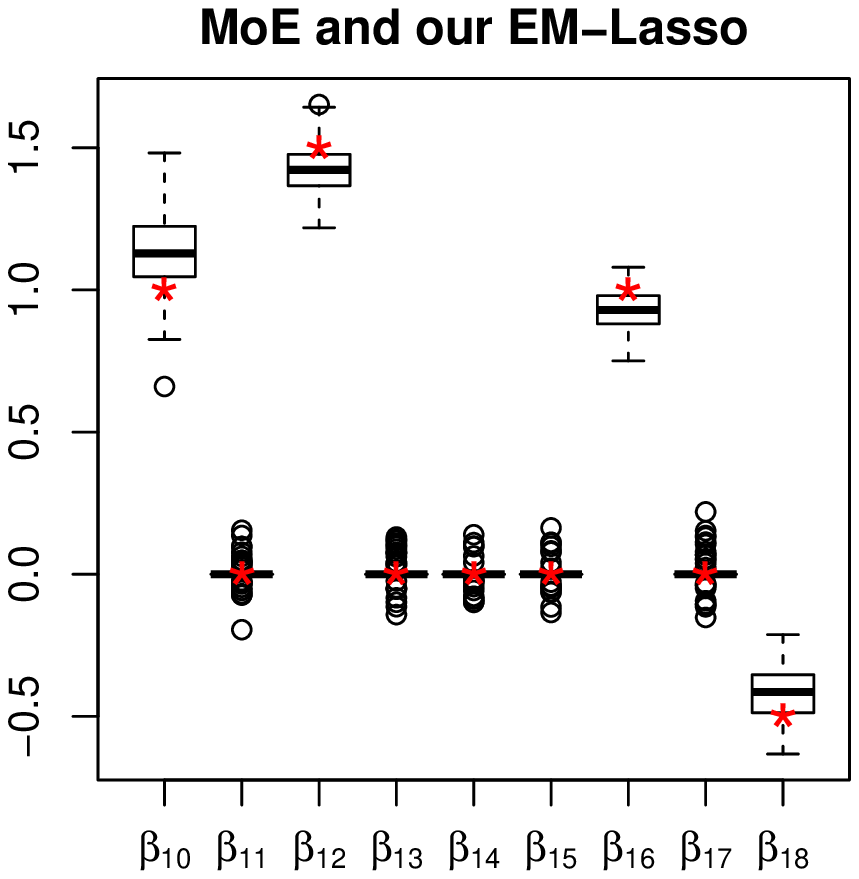}\\
    \includegraphics[width=0.42\textwidth]{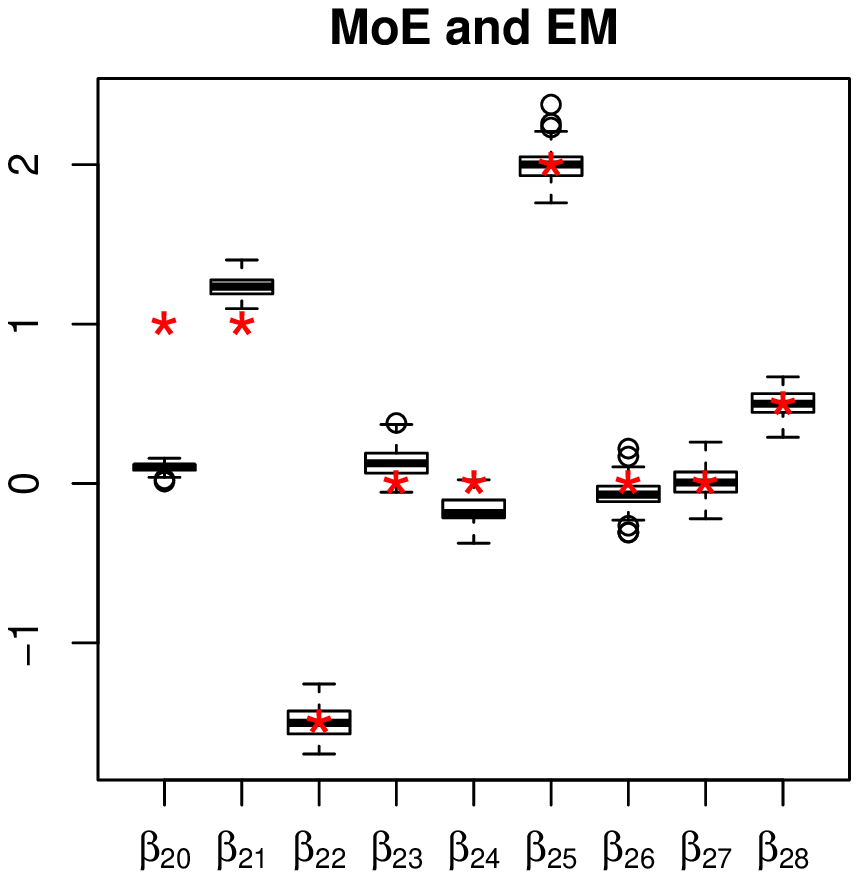}
    \includegraphics[width=0.42\textwidth]{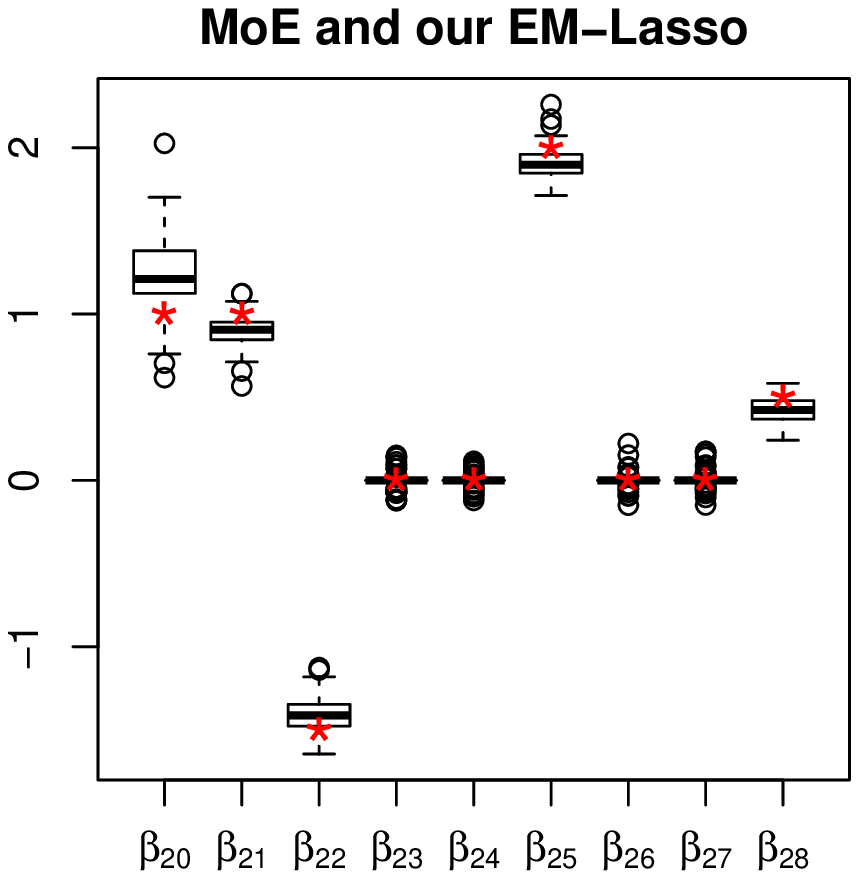}
    \caption{\label{fig:boxplot_experts}Boxplots of the estimated expert network parameters $\beta_{k,j}$: component $k=1$, top, and component $k=2$, bottom. The red stars are the true values.}
\end{figure}

\paragraph{Parameter estimation accuracy}Figure~\ref{fig:boxplot_gates} shows the estimated parameters for the gating network, with the error bars, for the proposed approach and for the standard MoGGE model. 
Similarly, Figure~\ref{fig:boxplot_experts} shows the estimated parameters of the gating network.
It can be seen on the two figures that, as expected, the proposed lasso-regularization approach with the proposed EM-Lasso algorithm, clearly provides models that are sparser, compared to the standard approach with EM, where the zero-coefficients are not precisely recovered. This is observed for both the gating function parameters, and the expert function parameters. While the penalized version we can see that it may be subject of a bias in estimating the non-zero coefficients, the parameter estimated and the bias are still reasonable. Hence, if one would to encourage sparsity, and to still have a good performance in density estimation, then the penalized MoGGE is a better choice, compared to the standard MLE of the MoGGE model. 

\paragraph{Sensitivity/specificity results} Table~\ref{tab:sensitivity_specificity} gives the sensitivity (S\textsubscript{1}) and specificity (S\textsubscript{2}) results for the two compared approaches. 
\begin{table}[h!]
    \centering
    \caption{Sensitivity (S\textsubscript{1}) and specificity (S\textsubscript{2}) results.}
    \label{tab:sensitivity_specificity}
    \begin{tabular}{|>{\centering\arraybackslash}p{3.5cm}|>{\centering\arraybackslash}p{1cm}|>{\centering\arraybackslash}p{1cm}|>{\centering\arraybackslash}p{1cm}|>{\centering\arraybackslash}p{1cm}|>{\centering\arraybackslash}p{1cm}|>{\centering\arraybackslash}p{1cm}|}\hline
        \multirow{2}{*}{Method} & \multicolumn{2}{c|}{Expert 1} & \multicolumn{2}{c|}{Expert 2} & \multicolumn{2}{c|}{Gate}\\\cline{2-7}
        & S\textsubscript{1} & S\textsubscript{2} & S\textsubscript{1} & S\textsubscript{2} & S\textsubscript{1} & S\textsubscript{2}\\\hline
    MoGGE-EM & 0.000 & 1.000 & 0.000 & 1.000 & 0.000 & 1.000\\\hline
    MoGGE-EMLasso-BIC & 0.790 & 1.000 & 0.785 & 1.000 & 0.779 & 1.000\\\hline
    \end{tabular}
\end{table}Note that here since we have two components, then only the estimation of one Gaussian gating function is considered, as the parameters of the other one are zeros. 
It can be seen that, none of the parameters in the non penalized model has a null value.
The penalized model provides naturally sparser models compared to the standard non-penalized one.

\paragraph{Clustering results}We calculate the accuracy of clustering for each data set. The results in terms of correct classification rate and ARI values are provided in Table~\ref{tab:accuracy}.
We can see that the classification rate as well as as the Adjusted Rand Index are very close for the two methods, with a slight advantage to the proposed approach.
\begin{table}[!h]
    \centering
    \caption{Clustering results: correct classification rate and Adjusted Rand Index.}
    \label{tab:accuracy}
    \begin{tabular}{|>{\centering\arraybackslash}p{3.5cm}|>{\centering\arraybackslash}p{2.5cm}|>{\centering\arraybackslash}p{2.5cm}|}\hline
        Model & C.rate & ARI\\\hline
        MoGGE - EM & 97.25\%\textsubscript{(0.8770\%)} & 89.28\%\textsubscript{(3.325\%)}\\\hline
        MoGGE-EMLasso-BIC & 97.43\%\textsubscript{(0.8521\%)} & 89.99\%\textsubscript{(3.231\%)}\\\hline
    \end{tabular}
\end{table}

\begin{figure}[!h]
    \centering
    \includegraphics[width=\textwidth]{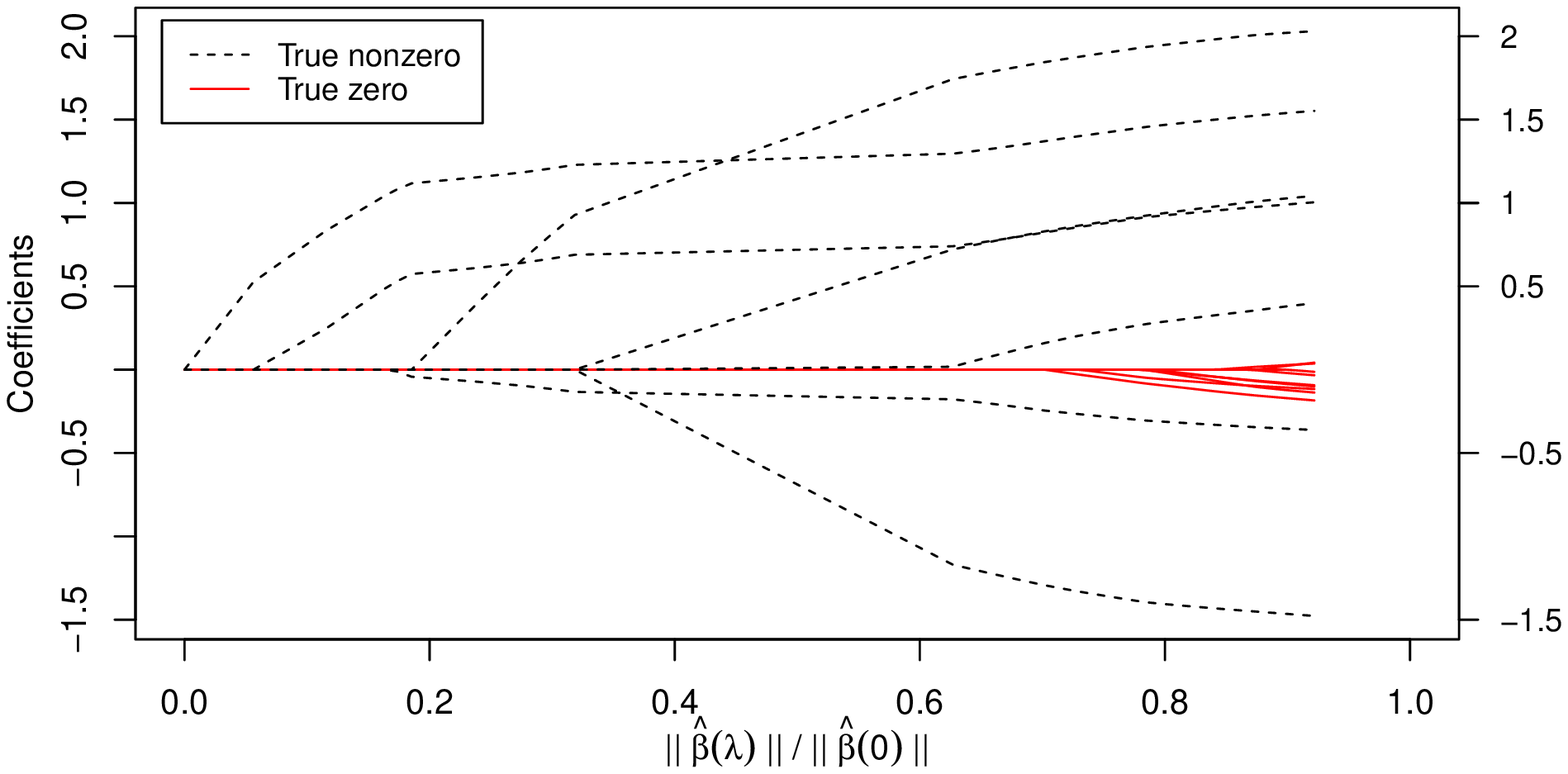}
    \includegraphics[width=\textwidth]{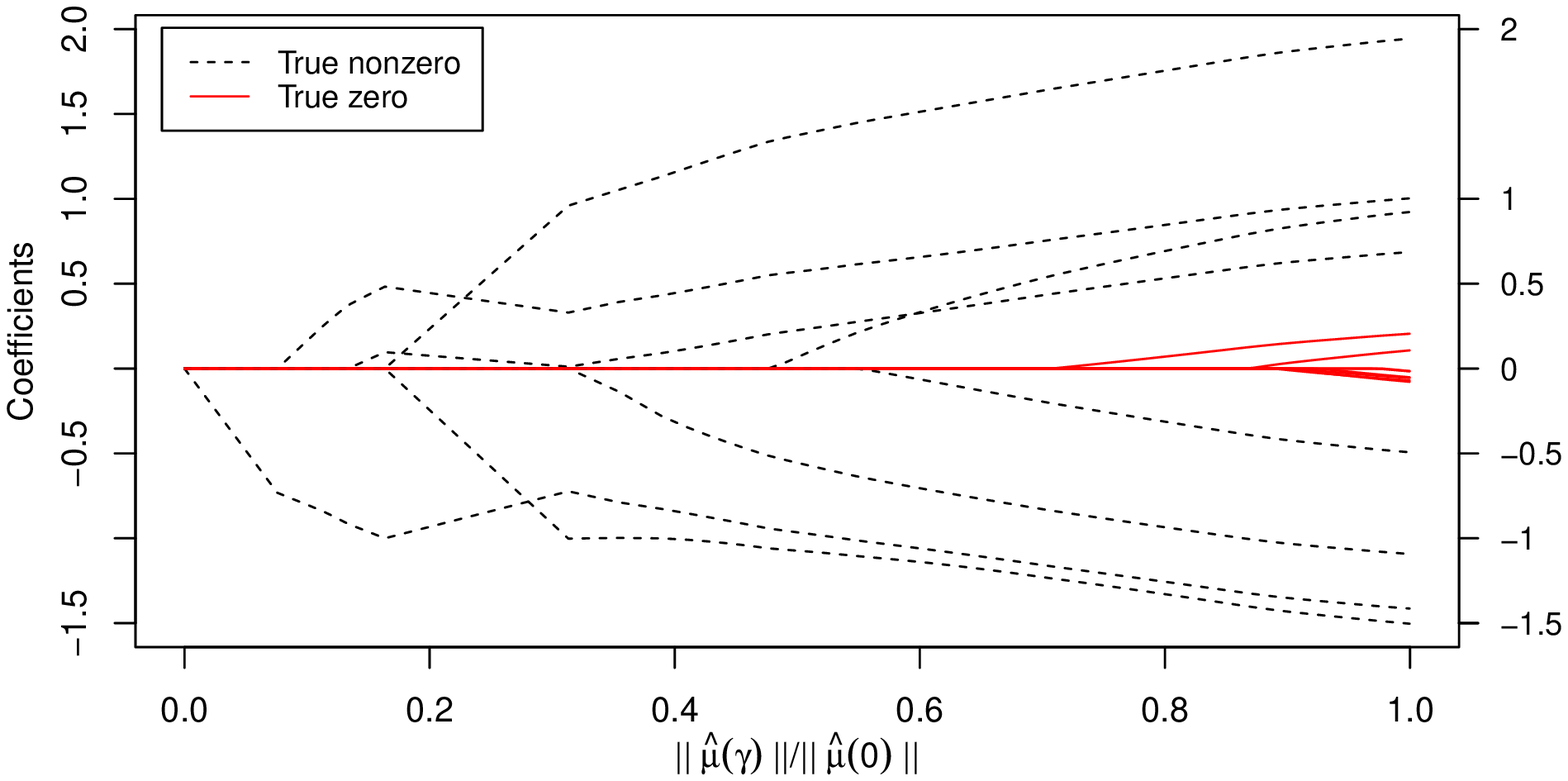}
    \caption{\label{fig:lasso_paths}Lasso paths of the estimated gating network parameters (top) and expert network parameters (bottom). The solid line represents the values of the true non-zero values, and the dashed line represents the true zero values.} 
\end{figure}

\paragraph{Selecting the sparsity tuning parameters} We compute the Lasso path for a sample with same parameters as presented at the beginning of the section.
On Figure~\ref{fig:lasso_paths}, we observe that even with very small values (null value as well, i.e. non penalized MoE) of $\gamma$, the true zero parameters have values very close to zero.
We also note that for values of ratio close to 0.8 for both $\lambda$ and $\gamma$, almost every true zero parameters have null values and the slight bias introduced in the true nonzero parameters is  reasonable.

\section{Conclusion and future work}
\vspace{-.3cm}
In this paper, the mixture of Gaussian-gated experts is studied towards modeling and clustering of heterogeneous regression data with high-dimensional predictors. 
A regularized MLE approach is proposed to simultaneously perform parameter estimation and feature selection. 
The developed EM-Lasso algorithm to fit the model relies on coordinate ascent updates of the regularized parameters, and its application in numerical experiments clearly shows it provides sparse models. Its performance is also compared to the state-of-the art fitting with the EM algorithm, shows its good performance, in particular in terms of sparsity. 
The diagonal hypothesis of the covariance matrix to derive the regularization \eqref{eq:pen joint loglik}
is now being relaxed, so that the regularization is on the elements of the precision matrix, i.e a graphical Lasso regularization. A future extension will also consider multivariate response with dedicated sparsity on the matrices of regression coefficients.

\section*{Acknowledgments}
\vspace{-.3cm}
This research is supported by Ethel Raybould Fellowship (Univ. of Queensland), ANR SMILES ANR-18-CE40-0014, and R{\'e}gion Normandie RIN AStERiCs.
 
\bibliographystyle{splncs04}
\vspace{-.4cm}
\bibliography{REFERENCES}

\end{document}

%% file: mymathsym.tex
\newcommand{\Bs}[1]{\boldsymbol{#1}}

\newcommand{\ba}{\mathbf{a}}

\newcommand{\bC}{\mathbf{C}}

\newcommand{\bB}{\mathbf{B}}

\newcommand{\bR}{\mathbf{R}}

\newcommand{\bw}{\mathbf{w}}
\newcommand{\bW}{\mathbf{W}}

\newcommand{\bX}{\mathbf{X}}

\newcommand{\bSigma}{\mathbf{\Sigma}}

\newcommand{\bsm}{\boldsymbol{m}}

\newcommand{\bsr}{\boldsymbol{r}}

\newcommand{\bsv}{\boldsymbol{v}}
\newcommand{\bsV}{\boldsymbol{V}}
\newcommand{\bsw}{\boldsymbol{w}}

\newcommand{\bsx}{\boldsymbol{x}}
\newcommand{\bsX}{\boldsymbol{X}}
\newcommand{\bsy}{\boldsymbol{y}}
\newcommand{\bsY}{\boldsymbol{Y}}

\newcommand{\cN}{\mathcal{N}}

\newcommand{\cL}{\mathcal{L}}

\newcommand{\cD}{\mathcal{D}}

\newcommand{\bsmu}{\boldsymbol{\mu}}

\newcommand{\bsbeta}{\boldsymbol{\beta}}

\newcommand{\bstheta}{\boldsymbol{\theta}}

\newcommand{\bsvPsi}{\boldsymbol{\varPsi}}

\newcommand{\bstau}{\boldsymbol{\tau}}

\newcommand{\E}{\mathbb{E}}

\newcommand{\Pro}{\mathbb{P}}
\newcommand{\R}{\mathbb{R}}